\documentclass[10pt,twocolumn]{article}

\usepackage[pagenumbers]{cvpr} 
\pdfoutput=1
\usepackage{times}
\usepackage{epsfig}
\usepackage{graphicx}
\usepackage{amsmath}
\usepackage{amssymb}
\usepackage{graphicx}
\usepackage{amsmath}
\usepackage{float}
\usepackage{amssymb}
\usepackage{booktabs}
\usepackage[utf8]{inputenc} 
\usepackage[T1]{fontenc}    
\usepackage{url}            
\usepackage{booktabs}       
\usepackage{amsfonts}       
\usepackage{nicefrac}       
\usepackage{microtype}      
\usepackage{xcolor}         
\usepackage{tabularx}
\usepackage{graphicx}
\usepackage{amsmath}
\usepackage{subcaption}
\usepackage{tabulary,multirow,overpic,xcolor}
\definecolor{mygray}{gray}{0.92}
\definecolor{baselinecolor}{gray}{.9}
\newcommand{\baseline}[1]{\cellcolor{baselinecolor}{#1}}
\usepackage{listings}
\usepackage{amsthm}
\usepackage{tabulary}
\usepackage{colortbl}
\usepackage{enumitem}
\usepackage{braket}
\usepackage{array}

\usepackage{float}
\usepackage{enumitem}
\usepackage{booktabs}
\usepackage{algorithm}
\usepackage{listings}
\usepackage{tabu}
\usepackage{pifont} 
\usepackage{marvosym}

\def\x{$\times$}

\newcolumntype{x}[1]{>{\centering\arraybackslash}p{#1pt}}
\newcolumntype{y}[1]{>{\raggedright\arraybackslash}p{#1pt}}
\newcolumntype{z}[1]{>{\raggedleft\arraybackslash}p{#1pt}}

\newlength\savewidth\newcommand\shline{\noalign{\global\savewidth\arrayrulewidth
		\global\arrayrulewidth 1pt}\hline\noalign{\global\arrayrulewidth\savewidth}}
\newcommand{\tablestyle}[2]{\setlength{\tabcolsep}{#1}\renewcommand{\arraystretch}{#2}\centering\footnotesize}

\definecolor{linkcol}{RGB}{233, 4, 141}

\definecolor{xycolor}{RGB}{60, 120, 216}
\definecolor{xycolor}{HTML}{0071bc}

\definecolor{wcolor}{RGB}{103, 78, 167}

\definecolor{dcolor}{RGB}{166, 77,21}

\definecolor{gcolor}{RGB}{204, 102, 153}

\definecolor{tcolor}{RGB}{34,139,34}

\definecolor{iterc}{RGB}{91,196,159}

\definecolor{epochc}{RGB}{96,172,252}

\definecolor{eicolor}{RGB}{153, 51, 102}

\definecolor{citecolor}{HTML}{0071BC}
\definecolor{linkcolor}{HTML}{ED1C24}

\usepackage{hyperref}
\hypersetup{breaklinks=true,letterpaper=true,colorlinks}

\usepackage[capitalize]{cleveref}
\crefname{section}{Sec.}{Secs.}
\Crefname{section}{Section}{Sections}
\Crefname{table}{Table}{Tables}
\crefname{table}{Tab.}{Tabs.}

\newcommand\blfootnote[1]{%
  \begingroup
  \renewcommand\thefootnote{}\footnote{#1}%
  \addtocounter{footnote}{-1}%
  \endgroup
}
\begin{document}

\title{MGMAE: Motion Guided Masking for Video Masked Autoencoding}

\author{Bingkun Huang\textsuperscript{1,2} \quad Zhiyu Zhao\textsuperscript{1,2} \quad Guozhen Zhang\textsuperscript{1} \quad Yu Qiao\textsuperscript{2} \quad Limin Wang\textsuperscript{1,2,~\Letter} \\ $^1$ State Key Laboratory for Novel Software Technology, Nanjing University \quad $^2$ Shanghai AI Lab \\ [0.2cm]
{\bf \url{https://github.com/MCG-NJU/MGMAE}} 
}

\maketitle

\begin{abstract}
Masked autoencoding has shown excellent performance on self-supervised video representation learning. Temporal redundancy has led to a high masking ratio and customized masking strategy in VideoMAE. In this paper, we aim to further improve the performance of video masked autoencoding by introducing a motion guided masking strategy. Our key insight is that motion is a general and unique prior in video, which should be taken into account during masked pre-training. Our motion guided masking explicitly incorporates motion information to build temporal consistent masking volume. Based on this masking volume, we can track the unmasked tokens in time and sample a set of temporal consistent cubes from videos. These temporal aligned unmasked tokens will further relieve the information leakage issue in time and encourage the MGMAE to learn more useful structure information. We implement our MGMAE with an online efficient optical flow estimator and backward masking map warping strategy. We perform experiments on the datasets of Something-Something V2 and Kinetics-400, demonstrating the superior performance of our MGMAE to the original VideoMAE. In addition, we provide the visualization analysis to illustrate that our MGMAE can sample temporal consistent cubes in a motion-adaptive manner for more effective video pre-training. \vspace{-2mm}
\end{abstract}
\blfootnote{\Letter: Corresponding author (lmwang@nju.edu.cn).}
\section{Introduction}
\label{sec:intro}

\begin{figure}[t]
    \centering
    \begin{subfigure}[t]{0.48\linewidth}
        \centering
        \includegraphics[width=\linewidth]{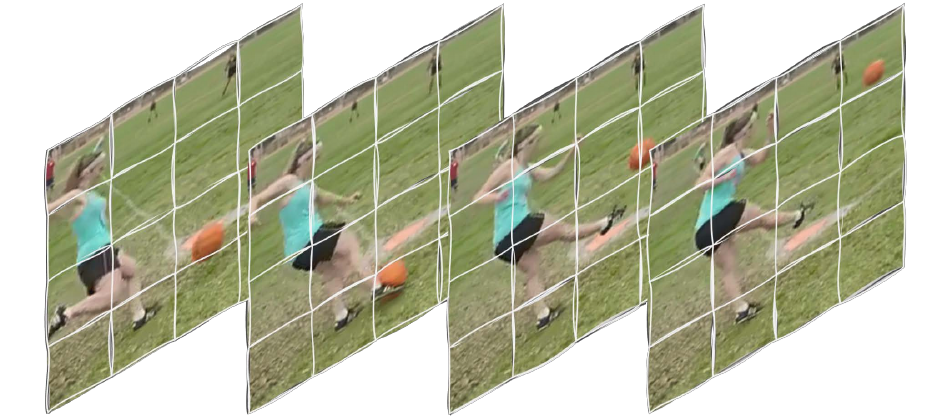}
        \caption{original video clip}
        \label{fig:orig_ball}
    \end{subfigure}
    \hfill
    \begin{subfigure}[t]{0.48\linewidth}
        \centering
        \includegraphics[width=\linewidth]{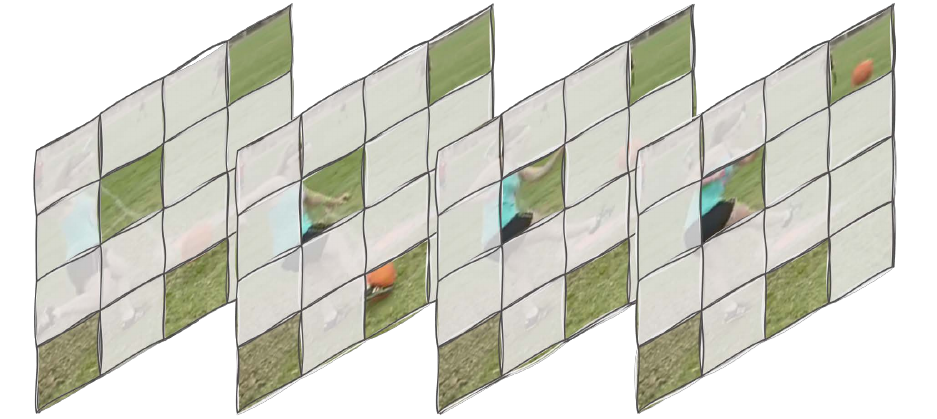}
        \caption{tube masking}
        \label{fig:tube_ball}
    \end{subfigure}
    \hfill
    \begin{subfigure}[t]{0.48\linewidth}
        \centering
        \includegraphics[width=\linewidth]{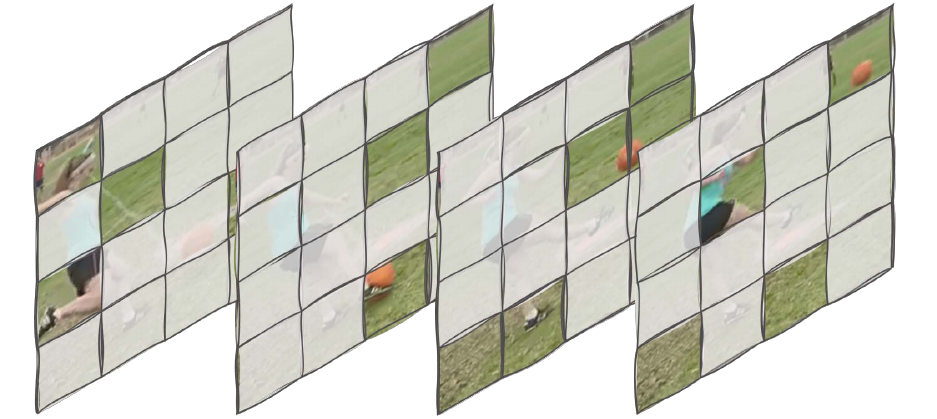}
        \caption{random masking}
        \label{fig:random_ball}
    \end{subfigure}
    \hfill
    \begin{subfigure}[t]{0.48\linewidth}
        \centering
        \includegraphics[width=\linewidth]{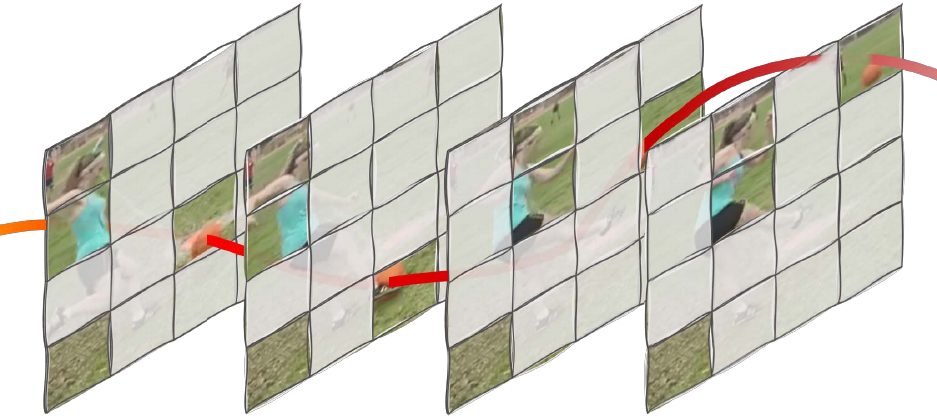}
        \caption{motion guided masking}
        \label{fig:mg_ball}
    \end{subfigure}
   \caption{{\bf Comparison of different masking strategies}. Masked autoencoding~\cite{devlin2019bert,he2021masked} has been explored in video domain for self-supervised pre-training by employing different masking strategies: random masking~\cite{feichtenhofer2022maest} and tube masking~\cite{tong2022videomae}. We propose to track masking maps under the guidance of motion information (termed as motion guided masking). Our resulting MGMAE can build a more challenging and meaningful task for video pre-training.}
   \label{fig:motivation}
   \vspace{-5mm}
\end{figure}

Attention-based Transformer~\cite{attention} has witnessed great success in computer vision since the introduction of Vision Transformer (ViT)~\cite{vit}. It has been applied for a variety of vision tasks and obtains state-of-the-art performance, such as image classification~\cite{deit,deepvit,t2t}, object detection~\cite{swin,pvt}, semantic segmentation~\cite{segformer}, and object tracking~\cite{mixformer}. Thanks to this high performance, ViT models have been also applied to the video domain for action recognition~\cite{timesformer,vivit} and detection~\cite{RTD,actionformer}. However, the high capacity of Transformer often demands pre-training on a large-scale dataset to reduce the over-fitting risk of subsequent fine-tuning. Therefore, an effective pretraining strategy of ViT is particularly important for obtaining excellent performance in the video domain due to the smaller video dataset.

The early video transformers~\cite{vivit,timesformer} often rely on the pre-training of image-based transformer derived from the large-scale image dataset~\cite{imagenet}. This pre-training scheme makes the learnt video model to be naturally biased by image-based ViTs. Recently, masked autoencoding (MAE)~\cite{tong2022videomae,feichtenhofer2022maest,wang2023videomaev2} has been explored for pre-training video transformer on the video dataset due to its simplicity and promising result in image domain~\cite{he2021masked}.
However, unlike the image, video data is equipped with an extra time dimension and exhibits the unique property of temporal redundancy and correlation. This property requires some customized designs on video masked autoencoder compared with image-based MAE. For example, VideoMAE and MAE-ST both propose to use an extremely high masking ratio in video masked autoencoder pre-training to improve its performance. In addition, VideoMAE devises a tube masking strategy of dropping tokens at the same position across frames to further relieve the information leakage in time. This tube masking approach, though straightforward, makes the assumption of no or small motion occurring between adjacent frames. Such an assumption might be not true for some scenarios with high-speed motion.

Based on the above analysis, in this paper, we aim to propose a new masking strategy for improving video masked autoencoder pre-training, by explicitly using motion information to reduce information leakage in time. Specifically, we devise the {\em Motion Guided Masking} in the video masked encoder processing and the resulted masked autoencoder is termed as {\bf MGMAE}. Motion is general prior information contained by video. The optical flow representation explicitly encodes the movement of each pixel from the current frame to the next one. We propose to use this optical flow to align masking maps between adjacent frames to build consistent masking volumes across time. The consistent masking volumes enable to build a more challenging reconstruction task by enforcing only a small set of cube tracks visible to the encoder. Hopefully, this motion guided masking can further relieve the risk of information leakage in time and encourage learning more meaningful visual representations.

More specifically, we use an online and lightweight optical flow estimator (RAFT~\cite{teed2020raft}) to capture motion information, which could be seamlessly integrated into the existing VideoMAE framework. To build the temporally consistent masking volume, we first randomly generate an initial masking map at the base frame. Then, we use the estimated optical flow to warp the initial masking map to adjacent frames. With multiple warping operations, we build the temporal consistent masking volume for all frames in the video. Finally, based on this masking volume, we sample a set of visible tokens to MAE encoders with top-k selection based on a frame-wise manner. The same autoencoding process with the original VideoMAE is applied to these sampled tokens for video pretraining. With this simple motion guided masking, we are able to further increase the difficulty of video pre-training task and thus lead to a better pre-trained model for subsequent fine-tuning. 

We mainly verify the effectiveness of the proposed MGMAE on the datasets of Something-Something V2~\cite{sth} and Kinetics-400~\cite{kinetics400} by comparing them with the original tube masking in VideoMAE. The results demonstrate that MGMAE pre-training can result in more powerful video foundation models with higher fine-tuning accuracy on the downstream tasks. In particular, on the motion-centric benchmark of Something-Something, the improvement of MGMAE is more evident, implying that our motion guided masking is adaptive to motion variations and can better capture temporal structure information for pre-training. We hope our findings can inspire some specific and unique designs in video masked autoencoding with respect to image counterparts.

\section{Related Work}

\paragraph{Masked Visual Modeling.} Masked autoencoder is a long-standing unsupervised learning framework in computer vision. The early work presented general form of denoising autoencoder~\cite{VincentLLBM10,VincentLBM08} for learning representation by reconstructing the clean signal from the noisy inputs. The other work~\cite{PathakKDDE16} also treated masking modeling as inpainting missing regions from the surrounding context by using convolutions. Inspired by the great success of masked language modeling~\cite{devlin2019bert}, some works also attempted to apply this pre-training paradigm to the vision domain for self-supervised pre-training. For example, iGPT~\cite{iGPT20} followed the GPT work~\cite{radford2019language} in NLP and processed a sequence of pixels for casual prediction of the next pixels. The original ViT~\cite{vit} used the masked token prediction as a self-supervised training step on large-scale image datasets but failed to obtain impressive results. Recently, several interesting works have obtained a great breakthrough in self-supervised image pre-training by using masked image modeling, such as BEiT~\cite{beit}, SimMIM~\cite{xie2021simmim}, and MAE~\cite{he2021masked}. BEiT~\cite{beit} directly followed the BERT framework and proposed to predict the discrete token label for masked patches, by requiring an explicit tokenizer to build the token dictionary. SimMIM~\cite{xie2021simmim} and MAE~\cite{he2021masked} shared the same design of directly predicting the pixels of masked patches without any tokenizer design. Furthermore, MAE~\cite{he2021masked} devised an asymmetric encoder-decoder architecture to speed up the masked image pre-training.

Since the great success in masked image modeling, some works have tried to extend this new pre-training paradigm to the video domain for self-supervised video pre-training. BEVT~\cite{wang2022bevt} and VIMPAC~\cite{vimpac} proposed to learn video representation by predicting discrete visual tokens in a similar way to BEiT. However, their performance improvement in video action recognition is limited. MaskFeat~\cite{wei2021masked} used the HOG features~\cite{hog} as the reconstructed targets of masked patches and achieved excellent performance on the video recognition with a multi-scale vision transformer. VideoMAE~\cite{tong2022videomae} and MAE-ST~\cite{feichtenhofer2022maest} extended the image MAE to the video domain for representation learning with vanilla vision transformer. They both proposed to use an extremely high masking ratio to deal with video data redundancy. Meanwhile, VideoMAE~\cite{tong2022videomae} used the tube masking to further increase the difficulty of reconstruction. Several works building upon VideoMAE have emerged. For instance, MAR~\cite{qing2023mar} reduced both training and inference costs by introducing running cell masking. Meanwhile, VideoMAE V2~\cite{wang2023videomaev2} proposes a dual masking strategy to decrease pre-training overhead, and by expanding both the model size and dataset, it further explores the scalability of VideoMAE. Our proposed motion guided masking aims to improve the performance of VideoMAE by building a more challenging masking and reconstruction task. In contrast to the original VideoMAE, our MGMAE explicitly use the optical flow to align the masking maps across frames and generate the temporal consistent masking volume to sample a set of visible tokens.

\begin{figure*}[t]
  \centering
   \includegraphics[width=0.95\linewidth]{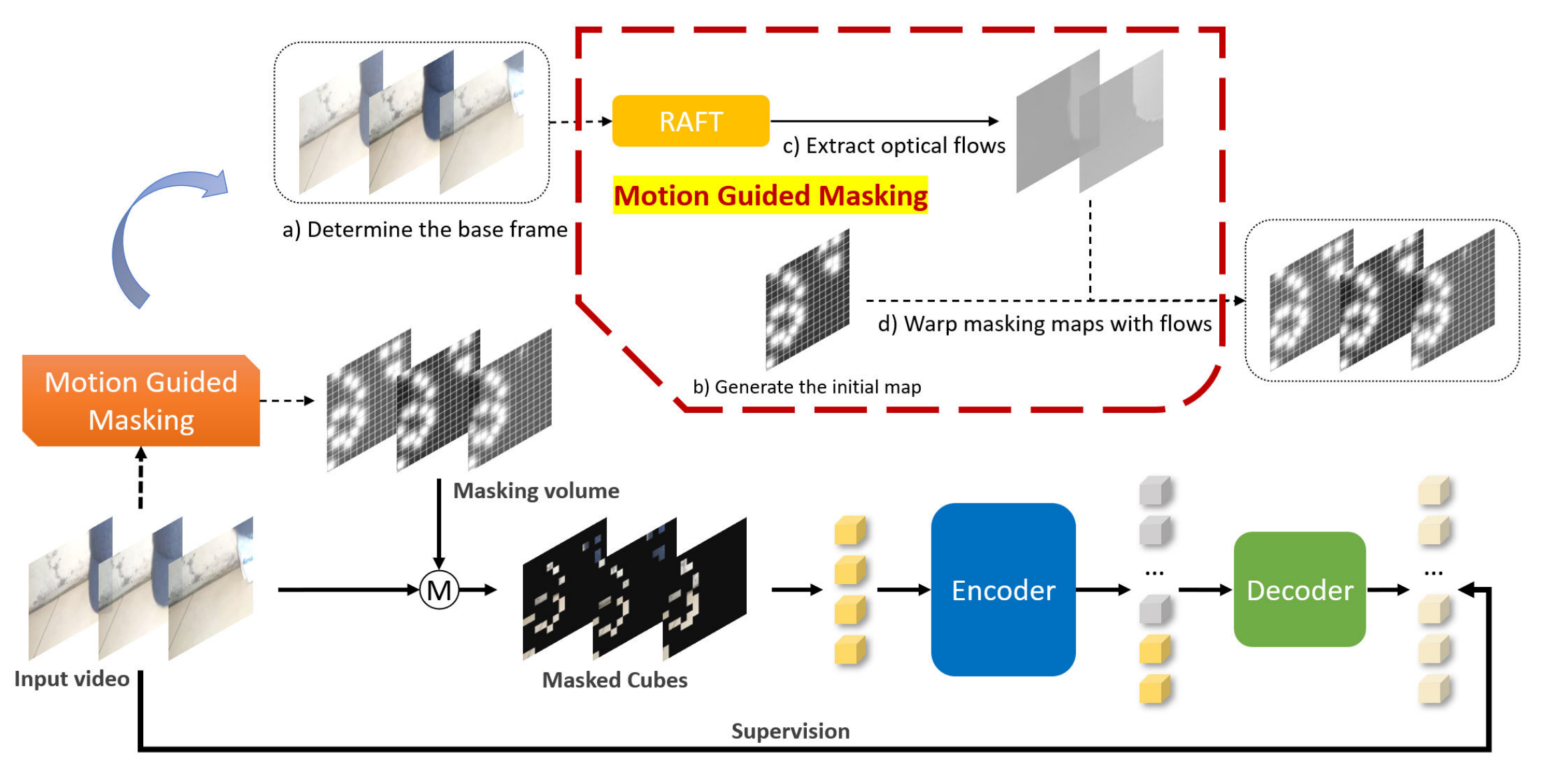}
   \caption{{\bf Pipeline of MGMAE}. Our MGMAE follows the simple pipeline of masking and reconstruction for video self-supervised pre-training. Our core design is to propose a motion guided masking strategy to generate temporal consistent masking volume. With this masking volume, we track the visible cubes and attain temporal consistency for masking maps. As a result, we can build a more challenging reconstruction task and encourage extracting more effective representations during masked self-supervised pretraining.}
   \label{fig:pipeline}
\end{figure*}

\paragraph{Motion Guided Modeling.} Motion information, such as optical flow, is a general prior information in videos and represents the unique characteristics distinct from images. Optical flow has been widely introduced to provide a strong prior in both low-level and high-level vision tasks on video.
For low-level video tasks, the motion is often used to align the information of auxiliary frames to the corresponding region of the target frame. In the case of video super-resolution, BasicVSR++~\cite{chan2022basicvsr++} uses optical flow to enhance the appearance of low-resolution frames by transferring features from neighbor frames. For video inpainting, Zhang~\etal~\cite{zhang2022flow} exploits the motion difference extracted by optical flows to instruct the attention retrieval in transformer for high-fidelity video inpainting.
As for video frame interpolation, mainstream methods leverage optical flow directly on the image to synthesize the intermediate frame, such as DAIN~\cite{bao2019depth} and RIFE~\cite{huang2020rife}, while Zhang~\etal~\cite{zhang2023extracting} introduces a unified operation utilizing inter-frame attention to concurrently extract motion and appearance information, and blends a hybrid CNN and Transformer design for efficiency and fine-grained detail preservation.
For high-level video tasks, the optical flow is directly used as a data modality as network input for action recognition~\cite{tsn_journal,simonyan2014two}. TDD~\cite{tdd} utilized motion trajectories to pool deep convolutional features for action recognition. Trajectory Convolution~\cite{tconv} incorporated the motion information into temporal convolutional kernel design.
MSNet~\cite{kwon2020motionsqueeze} proposes a pluggable MotionSqueeze module to generate motion information across frames. 
VideoMS~\cite{hwang2022videoms} generates mask
maps by calculating the feature difference after patch embedding, making an attempt at dynamically adjusting mask positions.
AdaMAE~\cite{bandara2023adamae} introduces an end-to-end trainable adaptive masking strategy for MAEs, leveraging an auxiliary sampling network to prioritize tokens from high spatiotemporal information regions.
Yang~\etal~\cite{yang2020hierarchical} leverage hierarchical motion information to improve the extracted video features.
MotionFormer~\cite{motionformer} employed the trajectory for attention computation in the video transformer. TEA~\cite{tea} and TDN~\cite{tdn} used RGB difference to approximate the motion information and incorporate this information into the video CNN backbone design. MGSampler~\cite{mgsampler} explored the motion information to select a subset of representative frames for efficient video action recognition.
Our MGMAE shares the same spirit with these motion guided modeling works. We focus on employing motion information as a cue to generate masking maps for masked video pre-training.

\section{Method}
In this section, we first revisit the pre-training paradigm of VideoMAE to well introduce our MGMAE in \cref{sec:revisit}. Then we present the details of motion guided masking map generation in \cref{sec:generate_map}. Finally, we describe the MGMAE pre-training under temporal consistent masking maps in \cref{sec:sample_token}.

\subsection{VideoMAE revisited}
\label{sec:revisit}
VideoMAE is a simple masked video autoencoder with an asymmetric encoder-decoder architecture with an extra cube embedding to handle the input sampled frames. Next, we briefly revisit its implementation detail.

\paragraph{Cube Embedding.} VideoMAE divides the input video clip $\mathbf{I}$ of size $T \times 3 \times H \times W$ into non-overlapping cubes $\mathbf{C} = \left\{ \mathbf{C}_i \mid \mathbf{C}_i \in \mathbb{R}^{2 \times 16 \times 16 \times 3} \right\}_{i=1}^N$, where $N = \frac{T}{2} \times \frac{H}{16} \times \frac{W}{16}$ is the number of cubes. Then apply cube embedding on the cubes to produce the video tokens $\mathbf{T} = \left\{ \mathbf{T}_i \mid \mathbf{T}_i \in \mathbb{R}^D \right\}_{i=1}^N$, where $T_i$ represents the cube embedding with positional encoding, and $D$ is the channel.

\paragraph{Masking Strategy.} VideoMAE uses the tube masking strategy with an extremely high masking ratio $\rho$ (\ie 90\%), which samples the same spatial positions across all frames of the input video clip. Specifically, VideoMAE first generate a $\frac{H}{16} \times \frac{W}{16}$ binary mask map $\mathbf{M}'$ where $0$ represents unmasked and $1$ represents masked. Then it replicates it in temporal dimension and then flattens it to produce the token-level mask map $\mathbf{M}$ whose size is $N$ for the input video clip. We denote $\mathbb{M}$ as the masking maps.

\paragraph{Encoder.} The encoder is a vanilla ViT with joint space-time attention \cite{timesformer}. For computation efficiency, only the unmasked visible tokens $\mathbf{T}^v = \left\{ \mathbf{T}_i \right\}_{i\notin \mathbb{M}}$ added with the fixed positional embedding are fed into the encoder to obtain the latent features $\mathbf{Z}$ of size $N_v \times D$, where $N_v = \lfloor (1-\rho) N\rfloor$ is the total number of the unmasked visible tokens.

\paragraph{Decoder.} The decoder is a narrower and shallower ViT than the encoder. It takes the concatenated token sequences as input, which is formed by the concatenation between the latent features $\mathbf{Z}$ and the learnable $\left[ \texttt{MASK} \right]$ tokens with the fixed position embedding added, to reconstruct the normalized video cubes $\hat{\mathbf{C}} = \left\{ \hat{\mathbf{C}}_i \mid \hat{\mathbf{C}}_i \in \mathbb{R}^{2 \times 16 \times 16 \times 3} \right\}_{i=1}^N$. 

\paragraph{Loss.} The pre-training object is to minimize the \textit{Mean Square Error} Loss between the normalized $\mathbf{C}$ and $\hat{\mathbf{C}}$ on the masked positions, \ie $\frac{1}{\rho N} \sum_{i \in \mathbb{M}} \left|\hat{\mathbf{C}}_i - \mathrm{norm}(\mathbf{C}_i)\right|^2$.

After the pre-training, the encoder will be used as the backbone network to fine-tune on the downstream tasks to obtain a specialized model.

\subsection{Motion Guided Masking Map}
\label{sec:generate_map}
Time is a unique characteristic of video and has different properties with the space dimension. When devising masked video autoencoder, we need to carefully take this extra time dimension into account and come up with a customized design. Information leakage in time is an important issue in masked video pre-training.  When information leakage occurs, the model can easily reconstruct the masked cubes based on the visible tokens of adjacent frames. In this case, it will greatly reduce the difficulty of the reconstruction task and lead to a pre-trained model with poor fine-tuning performance. A trivial solution to information leakage is to increase the masking ratio. VideoMAE~\cite{tong2022videomae} and MAE-ST~\cite{feichtenhofer2022maest} increase the masking ratio to 90\% to greatly increase the difficulty of reconstruction. In addition, VideoMAE makes the small motion assumption and adapts the tube masking strategy, which masks the same spatial position in all frames. However, this small motion prior is not always true for motion-dominated videos. A more reasonable approach is to keep each object in the video clip visible or invisible at all times. To achieve this goal, we propose the motion guided masking strategy to replace the tube masking strategy in VideoMAE. The strategy has two procedures: we first use the optical flow as guidance to generate temporally consistent masking volumes of the input video clip and then sample the unmasked visible tokens based on the temporal consistent masking volume. We will detail these two procedures in \cref{sec:generate_map} and \cref{sec:sample_token}.

In general, the procedure for generating the temporal consistent masking volumes has four steps as follows.

\begin{itemize}
    \item Step 1: Determine the base frame $\mathbf{I}_b$, where $b$ is the index of the base frame.
    \item Step 2: Randomly generate a pixel-level initial mask map $\mathbf{M}_b$ with size $H \times W$ as the mask map of $\mathbf{I}_b$.
    \item Step 3: Extract the dense flows $\mathbf{F}$ bidirectionally from the base frame $\mathbf{I}_b$ in the input video clip $\mathbf{I}$.
    \item Step 4: Warp the initial masking map $\mathbf{M}_b$ under the guidance of dense flows $\mathbf{F}$ and progressively build the temporal consistent masking volume $\mathbf{M}$ of size $T \times H \times W$.
\end{itemize}

\paragraph{Determine the base frame.}\label{para:determine_frame} 
By default, we choose the \textit{middle frame} as the base frame. In motion guided masking, we need to ensure all objects in the base frame remain consistently visible or invisible in all frames of the input clip. Note that objects may (dis)appear over time due to object or camera movement, and warping the masking map under optical flow can result in some holes due to pixels mapped out of bounds.
So the choice of the base frame may have an impact on the suppression of information leakage. We ablate the choice of the base frame in \cref{sec:ablation} and the middle frame is the optimal choice.

\paragraph{Generate the initial mask map.}\label{para:init_map} 
We initialize a pixel-level masking map for the base frame with the distribution of the Gaussian Mixture Model (GMM). We use the masking map to indicate the visible or invisible state of the cubes in the base frame. Previous masking strategies usually adapt a token-level binary initialization, \ie either all $0$ or all $1$ within each token of size $2 \times 16 \times 16$, which actually breaks the continuity of the object surface texture.

Specifically, we first randomly pick $\hat{N}_v = \lfloor (1 - \rho) \times \frac{H}{16} \times \frac{W}{16}\rfloor$ tokens whose centers are denoted with $c = \left\{\vec{c_i} : (c_{i1}, c_{i2})\right\}^{\hat{N}_v}_i$. Then we generate 2D Gaussian distributions $\mathcal{N}_i(c_i, \sigma^2)$ centered on the midpoint of each token, where $\sigma$ is the standard deviation and taken as the cube size $(16, 16)$. Thus we will obtain the mixed Gaussian distribution $\mathcal{P}(c, \sigma^2) = \sum_{i}^{\hat{N}_v} \mathcal{N}_i(\vec{c}_i, \sigma^2)$ corresponding to the base frame.
And the probability density function of the mixed Gaussian distribution is used to indicate the probability that the cube (token) is visible in the base frame.

\paragraph{Extract optical flows.}\label{para:extract_flow}
We use both online and offline alternative methods to extract optical flow. Online method adapts $\mathrm{RAFT}$ \cite{teed2020raft} (the small version) to estimate the flows of the input video clip. Offline method applies the traditional TVL1~\cite{zach2007tvl1} algorithm to extract the dense flows between all adjacent frames in advance. We perform the consistent crop-resize-rescale operations when reading flows. Online and offline methods achieve the similar results. More details see in \cref{sec:ablation}.

In practice, we only extract the flows $\mathbf{F}$ forward and backward from the base frame $\mathbf{I}_b$, i.e.
\begin{equation}
    \mathbf{F} = \left\{\mathbf{\upsilon}_{i\rightarrow i+1} \right\}_{i=1}^{b-1} \cup \left\{\mathbf{\upsilon}_{i\rightarrow i - 1} \right\}_{i=b+1}^{T},
\end{equation}
where the flow $\mathbf{\upsilon}_{i\rightarrow j}$ denotes the flow from $\mathbf{I}_i$ to $\mathbf{I}_j$.

\paragraph{Warp masking maps with flows.}\label{para:warp_flow} 
We utilize the method of \textit{backward warping} to generate the temporal consistent masking map of the video clip in a progressive manner. Forward warping $\phi_{F}$ and backward warping $\phi_{B}$ are two opposite patterns of warping. Both can be effectively employed to construct the masking volume from the initial masking map. Regrettably, forward warping suffers from hole or occlusion problems, that is, no flow vectors may pass to a certain pixel, or there may be multiple flow vectors passing to the same pixel. In contrast, backward warping maps the pixels of a given map one by one to individual locations. It's noteworthy that while backward warping doesn't escape from the issue of holes caused by mapping out of bounds, these holes are usually fewer than in forward warping and tend to occur at the boundaries of the map, thus causing less damage to the information distribution.
For the holes caused by backward warping, we fill them with the values of $\mathbf{M}_b$ at the same position to simulate the tube masking strategy.

Formally, given the flows $\mathbf{F}$ and the base frame masking volume $\mathbf{M}_b$, the mask map $\mathbf{M}_i$ of $\mathbf{I}_i$ can be constructed as
\begin{equation}
    \mathbf{M}_i =
    \begin{cases}
     \phi_B(\mathbf{M}_{i + 1}, \mathbf{\upsilon}_{i\rightarrow i+1}), & 1 \leq i < b \\
     \mathbf{M}_b, & i = b \\
     \phi_B(\mathbf{M}_{i - 1}, \mathbf{\upsilon}_{i\rightarrow i-1}), & b < i \leq T
    \end{cases}.
\end{equation}
Subsequently, the entire masking volume $\mathbf{M} = \left\{ \mathbf{M}_i \right\}_{i=1}^T$ of the video clip $\mathbf{I}$ can be constructed by backward warping the flows $\mathbf{F}$ bidirectionally, originating from the base frame mask map $\mathbf{M}_b$.

\subsection{Motion Guided MAE}
\label{sec:sample_token}
We build our MGMAE based on the above motion guided masking map. The temporal consistent masking volume indicates the probability that the corresponding position in the adjacent frame is visible under optical flow tracking. In order to suppress information leakage as much as possible, we sample the video tokens with the highest visible probability along the temporal dimension.
Specifically, we first perform average pooling with kernel size $2 \times 16 \times 16$ on the masking volume $\mathbf{M}$ to obtain the token-level masking volume $\mathbf{M}'$ of size $\frac{T}{2} \times \frac{H}{16} \times \frac{W}{16}$. Then we pick the top-$\hat{N}_v$ locations of each mask map of size $\frac{H}{16} \times \frac{W}{16}$ along the temporal dimension and thus sample $N_v$ corresponding video tokens as the unmasked visible tokens.

These sampled tokens according to our temporal consistent masking volume are fed into the asymmetric encoder-decoder for autoencoding based pre-training. The resulted pre-training framework is called {\em Motion Guided Masked Autoencoder} (MGMAE). The pre-trained model by our MGMAE is applied in the same way with the original VideoMAE for fine-tuning the downstream tasks.

\paragraph{Discussion.} Previous works~\cite{tong2022videomae, feichtenhofer2022maest} extended MAE to the video domain. They opted for the random (agnostic) masking and tube (space-only) masking strategies, respectively. {\em Random masking} introduces no explicit inductive bias about the video's space and time dimension. It aims to present a unified feature representation learning framework with minimal domain knowledge. We argue that although this idea is simple, time is inherently a distinctive dimension from space. By recognizing this, we could better leverage this prior information for enhanced video masked autoencoding. {\em Tube masking} assumes that a large area of the frame contains no or small motion, and thus masking the same position across frames could greatly reduce the information leakage risk. However, for motion-dominated video datasets such as Something-Something, this assumption will no longer hold true. Our proposed {\em motion guided masking} offers a more general and conceptually simple solution to take temporal correlation into account. It could be viewed as an adaptive video masking strategy and create more challenging yet meaningful tasks in video pre-training. 

\begin{table*}[t!]
    \centering
    \small
    \begin{subtable}[t]{0.3\textwidth}
        \centering
        \begin{tabular*}{0.9\textwidth}{@{\extracolsep{\fill}}ccc}
        case & Acc@1 & Acc@5 \\\hline
        first frame & 70.5 & 92.7 \\
        random frame & 70.6 & 92.9 \\
        \baseline{middle frame} & \textbf{71.0} & \textbf{93.1}
        \end{tabular*}
        \caption{{\bf Base frame selection.} We perform ablation study to select the base frame as the first, random or the middle frame.}
        \label{tab:base_frame}
    \end{subtable}
    \hfill
    \begin{subtable}[t]{0.3\textwidth}
        \centering
        \begin{tabular*}{0.9\textwidth}{@{\extracolsep{\fill}}ccc}
        case & Acc@1 & Acc@5 \\\hline
        forward & 70.5 & 92.9 \\
        \baseline{backward} & \textbf{71.0} & \textbf{93.1}
        \end{tabular*}
        \caption{{\bf Warping method.} We choose forward or backward warping method for aligning masking maps.}
        \label{tab:warp_flow}
    \end{subtable}
    \hfill
    \begin{subtable}[t]{0.3\textwidth}
        \centering
        \begin{tabular*}{0.9\textwidth}{@{\extracolsep{\fill}}ccc}
        case & Acc@1 & Acc@5\\\hline
        clip-level & 70.7 & 92.9 \\
        \baseline{frame-level} & \textbf{71.0} & \textbf{93.1}
        \end{tabular*}
        \caption{{\bf Sampling strategy.} We perform a study to choose the visible token generation strategy based on motion guided masking volume.}
        \label{tab:sample_token}
    \end{subtable}
    \hfill
    \begin{subtable}[t]{0.3\textwidth}
        \centering
        \begin{tabular*}{0.9\textwidth}{@{\extracolsep{\fill}}ccc}
        case & Acc@1 & Acc@5 \\\hline
        token rand. & 70.9 & 93.0 \\
        pixel rand. & 70.8 & 93.0 \\
        \baseline{mixed Gauss} & \textbf{71.0} & \textbf{93.1}
        \end{tabular*}
        \caption{{\bf Masking initialization.} We compare three methods to generate the masking map in the base frame. Two binary random methods and one mixed Gaussian method.}
        \label{tab:init_method}
    \end{subtable}
    \hfill
    \begin{subtable}[t]{0.3\textwidth}
        \centering
        \begin{tabular*}{0.9\textwidth}{@{\extracolsep{\fill}}ccc}
        case & Acc@1 & Acc@5 \\\hline
        random & 70.8 & 92.9 \\
        invisible & 70.7 & 92.8 \\
        visible & 70.8 & \textbf{93.1} \\
        previous map & 70.6 & 92.8 \\
        \baseline{tube} & \textbf{71.0} & \textbf{93.1}
        \end{tabular*}
        \caption{{\bf Hole filling.}  We choose some baseline choices to fill the value in the hole place caused by warping. We also use the tube filling consistent with the tube masking.}
        \label{tab:fill_hole}
    \end{subtable}
    \hfill
    \begin{subtable}[t]{0.3\textwidth}
        \centering
        \begin{tabular*}{0.9\textwidth}{@{\extracolsep{\fill}}cccc}
        method & time & Acc@1 \\\hline
        None (VideoMAE) & 32 h & 69.6 \\
        TVL1 (offline) & 41 h & 71.2  \\
        RAFT with 6 iters & 43 h & \textbf{71.3} \\
        \baseline{RAFT with 12 iters} & 56 h & 71.0  \\
       \end{tabular*}
       \caption{{\bf Method of optical flow estimation.} We perform the ablation study to investigate the influence of different methods of optical flow estimation.}
       \label{tab:extract_flow}
    \end{subtable}
    \caption{Parts of the ablation experiments on the Something-Something V2 dataset. Our MGMAE pre-training is implemented with the 16-frame vanilla ViT-B backbone. All models are pre-trained for 800 epochs and the masking ratio is $\rho=90\%$. The inference protocol is to report the fine-tuning action recognition accuracy with 2 clips $\times$ 3 crops. The default choice for our model is colored in \colorbox{baselinecolor}{gray}. Although the default setting is not optimal in terms of the method of optical flow estimation, we believe that this does not affect the conclusions of the ablation experiments.}
    \label{tab:ablation}
\end{table*}


\section{Experiments}
\label{sec:exp}

\subsection{Dataset}
\label{sec:dataset}
Following the original VideoMAE, we evaluate our MGMAE on Kinetics-400 (K400) \cite{kinetics400} and Something-Something V2 (SSV2) \cite{sth}. K400 contains about $240$k training videos and $20$k validation videos from YouTube and the actions in K400 are usually coupled with specific objects or scenes, such as brushing teeth and playing piano. While SSV2 contains about $169$k training videos and $25$k validation videos, and the categories in SSV2 only care about specific motion patterns (e.g. push, pull). We first pre-train the video transformer with our MGMAE on the corresponding dataset for self-supervised representation learning. Then, we report the fine-tuning performance of pre-trained models on the target datasets for action recognition. In our MGMAE pre-training, we generally follow the setting and implementation of the original VideoMAE~\cite{tong2022videomae}. We use the RAFT~\cite{teed2020raft} to extract optical flow due to its efficiency and accuracy in our MGMAE pre-training.

\subsection{Ablation Studies}
\label{sec:ablation}
In this subsection, we conduct in-depth ablation experiments on the choice in each step of our motion guided masking strategy. We pre-train the ViT-base model 800 epochs on the SSV2 dataset with 16 80G-A100 GPUs, and then fine-tune the encoder on the SSV2 dataset for action recognition. All models share the same training schedule and report the $2 \text{ clips} \times 3 \text{ crops}$ accuracy.

\paragraph{Choice of the base frame.} In this study, we investigate the influence of base frame selection for initial masking generation process.
We compare the middle frame as the based frame with either the first or a random frame, and the result is shown in~\cref{tab:base_frame}. It implies the middle frame is the best.
\paragraph{Warping method with optical flow.} We compare two kinds of warping methods to align masking maps across frames as explained in~\cref{sec:generate_map}. As previously mentioned above, the forward warping often leads to more severe occlusion and hole problems in the masking warping process. On the contrary, backward warping can effectively relieve this issue and ensure a more smooth masking warping. The result in~\cref{tab:warp_flow} demonstrates that back warping contributes to better performance.

\paragraph{Sampling strategy of top-k visible tokens.} We examine and compare the two sampling strategies to select the visible tokens based on our temporal consistent masking volume. \emph{Frame-level} strategy samples the top-k locations for each frame independently, while \emph{clip-level} strategy samples the top-k locations for the entire video jointly. As in \cref{tab:sample_token}, the frame-level top-k sampling strategy achieves slightly better performance.

\paragraph{Masking initialization at the base frame.} We ablate the choice of generating the initial masking at the base frame as shown in \cref{tab:init_method}. The token-level initialization method divides the mask map into $\frac{H}{16} \times \frac{W}{16}$ tokens of size $16 \times 16$, and randomly sets 90\% tokens to 0 (representing masked) and 10\% tokens to 1 (representing unmasked). The pixel-level initialization method randomly sets 90\% pixels to 0 and 10\% pixels to 1. The initialization process of the mixed Gaussian method has been detailed in~\cref{sec:generate_map}. The result demonstrates that the mixed Gaussian initialization method works the best.

\paragraph{Hole filling method.} We investigate the various methods to handle the holes problem brought by the mapping out of bound in backward warping in \cref{tab:fill_hole}. To determine the real holes caused by warping, we set the $0$ in the initial mask map to value $1e-8$, and then the locations equal to $0$ in new mask maps are treated as the holes. We experiment with 5 methods to fill the holes: \emph{Invisible} method fills all holes with $0$, while \emph{Visible} method fills all holes with $1$. \emph{Random} method randomly fills the holes to $0$ with probability of masking ratio $\rho$ and to $1$ with probability of $1 - \rho$. \emph{Previous map} method fills holes using the values from the same spatial positions as the last generated mask map. Conversely, \emph{Tube} method fills the holes with the value from the corresponding positions of the initial mask map, aligning with tube masking principles. We see that the tube method performs the best among all the methods.

\paragraph{Method of optical flow estimation.} We evaluate the effect of different methods of optical flow estimation as shown in \cref{tab:extract_flow}. For the offline method, we use the TVL1 algorithm to extract optical flows in advance and it achieve a comparable accuracy to the online RAFT optical flow.
Although VideoMAE is 1.3 times faster to train than MGMAE with RAFT-small (set to 6 testing iterations) to estimate flows, MGMAE has a clear advantage in terms of performance and reducing the risk of overfitting.
We find that the offline method is not much faster than the online method because IO (reading optical flow from disk) will be a bottleneck to increase the training speed. Note that our default setting is not optimal, but the conclusions drawn in other ablation experiments should not be impacted.

\begin{figure}[t]
    \centering
    \includegraphics[width=\linewidth]{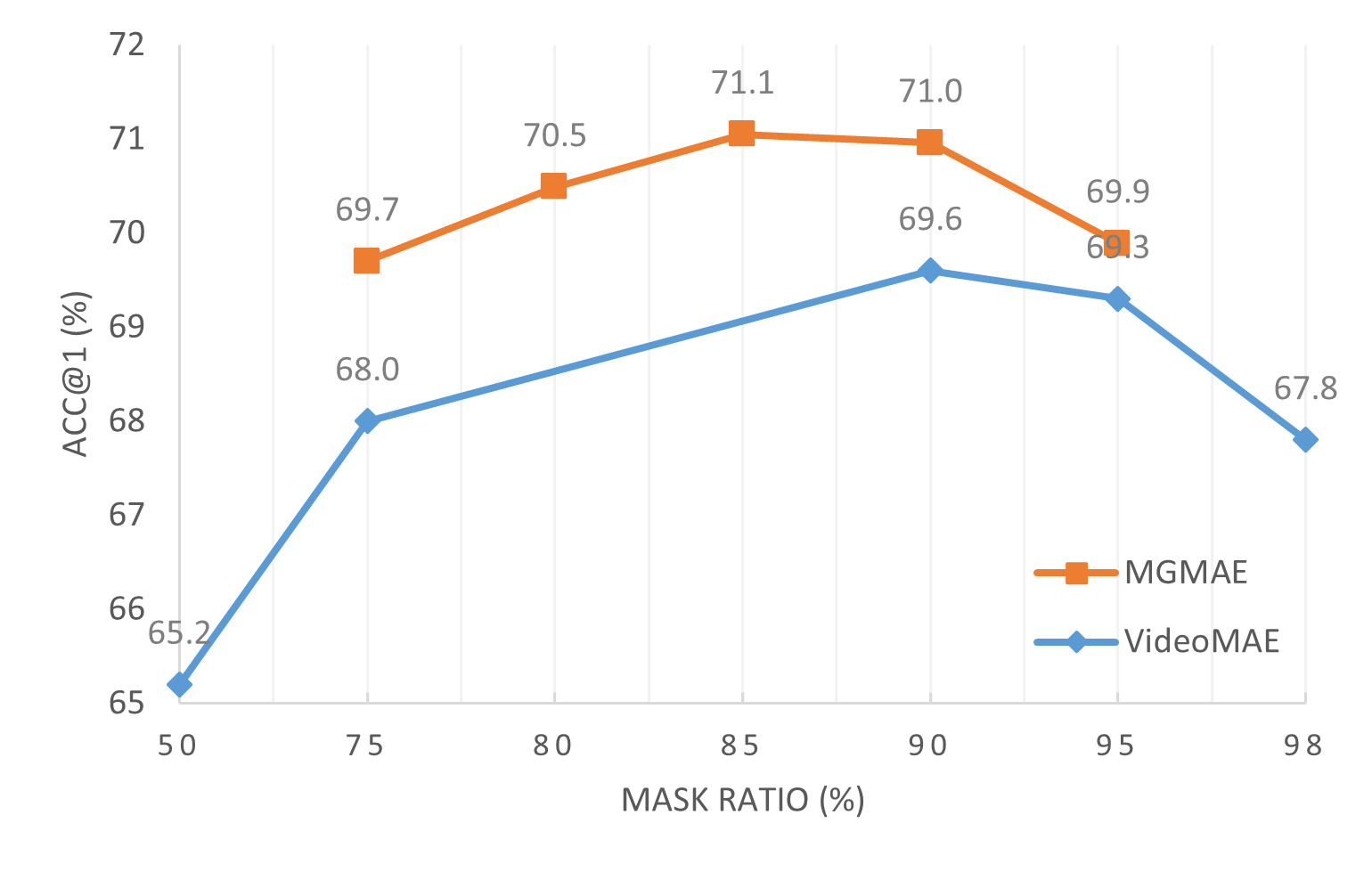}
    \caption{The effect of masking ratio on SSV2.}
    \label{fig:mask_ratio}
\end{figure}

\paragraph{Masking ratio.} The performance of MGMAE highlights the importance of improving masking strategies even at high masking rates (\eg 90\%). However, after applying MGMAE, it remains questionable whether such high masking rates are still necessary. Indeed, as pointed out by \cite{tong2022videomae, feichtenhofer2022maest}, blindly increasing the masking rate could potentially degrade the model performance. Our ablation study presented in \cref{fig:mask_ratio} shows that sustaining a extremely high masking rate of over 80\% is also crucial even with MGMAE. We think the video background and large objects mainly drive the need for a high masking ratio. Video backgrounds are often wide and simple. If the mask ratio is not high enough, the model can still rebuild pixels from other background parts, even if nearby frames mask similar sections. For large objects, a lower ratio might let the model use the texture from a different part of the object when another section is masked. It can also be observed that MGMAE performs optimally with 85\% masking ratio, but 90\% still seems to be a decent choice when considering the trade-off between training efficiency and performance.

\paragraph{Exposure of masked objects.} Another proposition worth considering is whether occasional exposure of masked objects help with masked video modeling pre-train. We completed a complementary experiment. Specifically, after building the masking volume, we add Gaussian noise on the mask map of \textit{one randomly selected frame}. This modification may provide a chance for masked objects to be exposed. The results showed an accuracy of 71.2\%, slightly higher than the default setting of 71.0\%.

\subsection{Main Results and Visualization Analysis}

After the detailed ablation study on the design of MGMAE, we further perform a deeper analysis by comparing it with the original VideoMAE. We also provide some intermediate visualization results to illustrate the motion guided sampling process.
\begin{table}[t!]
\small
    \centering
    \begin{tabular}{cc|cc}
    Model & Epochs & K400 & SSV2  \\
    \shline\hline
    VideoMAE & \multirow{2}{*}{800 / 800} & 0.5875 & 0.5278 \\
    MGMAE & & 0.6462 & 0.5820 \\
    $\Delta$ loss & - & $\Delta + 0.0605$ & $\Delta + 0.0542$ \\
    \hline
    VideoMAE & \multirow{2}{*}{1600 / 2400} & 0.5809 & 0.5122 \\
    MGMAE & & 0.6378 & 0.5659 \\
    $\Delta$ loss & - & $\Delta + 0.0569$ & $\Delta + 0.0537$
    \end{tabular}
    \caption{Pre-train loss comparison of MGMAE and VideoMAE.}
    \label{tab:pretrain_loss}
\end{table}
\begin{table}[t!]
\small
    \centering
    \begin{tabular}{cc|cc}
    Model & Epochs & K400 & SSV2  \\
    \shline\hline
    VideoMAE & \multirow{2}{*}{800 / 800} & 80.0 & 69.6 \\
    MGMAE & & 81.2 & 71.0 \\
    $\Delta$ Acc@1 & - & $\Delta + 1.2\%$ & $\Delta +1.4\%$ \\
    \hline
    VideoMAE & \multirow{2}{*}{1600 / 2400} & 81.5 & 70.8 \\
    MGMAE & & 81.8 & 72.3 \\
    $\Delta$ Acc@1 & - & $\Delta +0.3\%$ & $\Delta + 1.5\%$
    \end{tabular}
    \caption{Accuracy comparison of MGMAE and VideoMAE.}
    \label{tab:performance_compare}
\end{table}
\begin{figure}[t!]
    \centering
    \includegraphics[width=\linewidth]{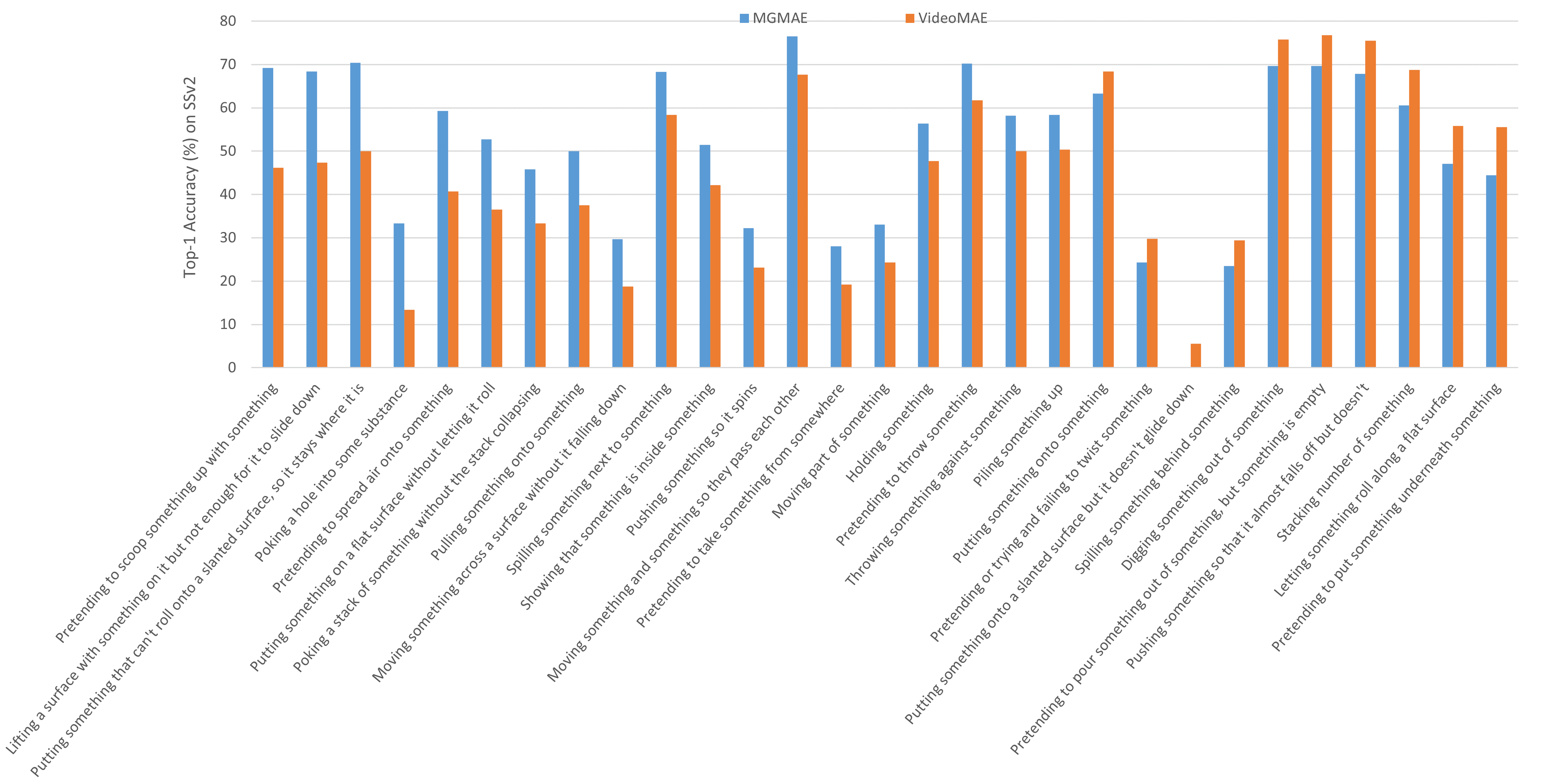}
    \caption{Comparative Accuracy of MGMAE and VideoMAE by Class.}
    \label{fig:acc_vs}
\end{figure}
\begin{table*}[t!]
\centering
\tablestyle{2.0pt}{1.04}
\begin{tabular}{l|c|c|c|c|c|c|c}
\textbf{Method} & \textbf{Backbone} & \textbf{Pre-train data} & \textbf{Frames} & \textbf{GFLOPs}  & \textbf{Param} & \textbf{Acc@1}  & \textbf{Acc@5} \\
\shline\hline
TEINet$_{En}$~\cite{teinet} &  \footnotesize{ResNet50$_{\times 2}$} &  \multirow{3}{*}{ImageNet-1K} & 8+16 &  99\x10\x3 & 50 & 66.5 & N/A \\
TANet$_{En}$~\cite{tanet} &  \footnotesize{ResNet50$_{\times 2}$} & & 8+16 & 99\x2\x3 & 51 & 66.0 & 90.1 \\
TDN$_{En}$~\cite{tdn} & \footnotesize{ResNet101$_{\times 2}$} &   &  8+16 & 198\x1\x3 & 88 & 69.6 & 92.2   \\
\hline
SlowFast~\cite{slowfast} &  ResNet101 &  \multirow{2}{*}{Kinetics-400} & 8+32 & 106\x1\x3 & 53 & 63.1 & 87.6 \\
MViTv1~\cite{mvit} & MViTv1-B &  &  64 & 455\x1\x3 & 37 & 67.7 & 90.9  \\
\hline
TimeSformer~\cite{timesformer} & ViT-B & \multirow{2}{*}{ImageNet-21K} &  8 & 196\x1\x3 & 121 & 59.5 &  N/A \\
TimeSformer~\cite{timesformer} & ViT-L &  &  64  & 5549\x1\x3 & 430 & 62.4 & N/A \\
\hline
ViViT FE~\cite{vivit} & ViT-L & \multirow{4}{*}{\footnotesize IN-21K+K400} &  32 & 995\x 4\x3 & N/A & 65.9 & 89.9 \\
Motionformer~\cite{motionformer} & ViT-B &     & 16 & 370\x1\x3 & 109  & 66.5 & 90.1 \\
Motionformer~\cite{motionformer} & ViT-L &     & 32 & 1185\x1\x3 & 382  & 68.1 & 91.2 \\
Video Swin~\cite{liu2021video}  & Swin-B &    & 32 & 321\x1\x3 & 88 & 69.6 & 92.7  \\
\hline
VIMPAC~\cite{vimpac} & ViT-L &HowTo100M+DALLE w/o label &  10 & N/A\x10\x3 & 307 & 68.1 & N/A \\
BEVT~ \cite{wang2022bevt} & Swin-B & IN-1K+K400+DALLE w/o label & 32 & 321\x1\x3 & 88 & 70.6 & N/A  \\
\hline
\textcolor{gray}{MAE-ST$_{1600e}$~\cite{feichtenhofer2022maest}} & \textcolor{gray}{ViT-L} & \textcolor{gray}{Kinetics-400 w/o label} & \textcolor{gray}{16} & \textcolor{gray}{597\x1\x3} & \textcolor{gray}{305} & \textcolor{gray}{72.1} & \textcolor{gray}{93.9} \\
\textcolor{gray}{MaskFeat{\scriptsize\textuparrow312}~\cite{wei2021masked}} & \textcolor{gray}{MViT-L} & \textcolor{gray}{Kinetics-600} &  \textcolor{gray}{40}  & \textcolor{gray}{2828\x1\x3} & \textcolor{gray}{218} & \textcolor{gray}{75.0} & \textcolor{gray}{95.0} \\
\hline
VideoMAE$_{800e}$~\cite{tong2022videomae} & ViT-B & \multirow{2}{*}{SSV2 w/o label} & 16 & 180\x2\x3 & 87 & 69.6 &  92.0 \\
VideoMAE$_{2400e}$~\cite{tong2022videomae} & ViT-B &  & 16 & 180\x2\x3 & 87 & 70.8 & 92.4 \\
\shline\hline
MGMAE$_{800e}$ & ViT-B & \multirow{2}{*}{SSV2 w/o label} & 16 & 180\x2\x3 & 87 & 71.0 & 93.1 \\
\textbf{MGMAE$_{2400e}$} & ViT-B & & 16 & 180\x2\x3 & 87 & \textbf{72.3} & \textbf{93.5}
\end{tabular}
\vspace{-2mm}
\caption{\textbf{Comparison on the Something-Something V2 dataset}. We only list the results obtained with the similar backbones.}
\label{tab:sota-ssv2}
\vspace{-1mm}
\end{table*}

\begin{table*}[t!]
\centering
\tablestyle{2.0pt}{1.04}
\begin{tabular}{l|c|c|c|c|c|c|c}
\textbf{Method} & \textbf{Backbone} & \textbf{Pre-train data} & \textbf{Frames} & \textbf{GFLOPs}  & \textbf{Param} & \textbf{Acc@1}  & \textbf{Acc@5} \\
\shline\hline
NL I3D~\cite{nonlocal} & ResNet101 &   \multirow{3}{*}{ImageNet-1K}  & 128 & 359\x10\x3  & 62  & 77.3 & 93.3    \\
TANet~\cite{tanet} &  ResNet152 & & 16 & 242\x4\x3 & 59 & 79.3 & 94.1 \\
TDN$_{En}$~\cite{tdn} & ResNet101 &   & 8+16 & 198\x10\x3 & 88 & 79.4 & 94.4   \\
\hline
TimeSformer~\cite{timesformer} & ViT-L & \multirow{4}{*}{ImageNet-21K}   & 96 & 8353\x1\x3 & 430 & 80.7 & 94.7 \\
ViViT FE~\cite{vivit} & ViT-L &  & 128 & 3980\x 1\x3 & N/A & 81.7 & 93.8 \\
Motionformer~\cite{motionformer} & ViT-L &    & 32 & 1185\x10\x3 & 382  & 80.2 & 94.8 \\
Video Swin~\cite{liu2021video}  & Swin-B &   & 32 & 282\x4\x3 & 88 & \textbf{82.7} & \textbf{95.5}  \\
\hline
VIMPAC~\cite{vimpac} & ViT-L & \scriptsize{HowTo100M+DALLE w/o label} & 10 & N/A\x10\x3 & 307 & 77.4 & N/A \\
BEVT~\cite{wang2022bevt}  & Swin-B & IN-1K+DALLE w/o label & 32 & 282\x4\x3 & 88 & 80.6 & N/A  \\
\hline
ip-CSN~\cite{csn} & ResNet152 & \multirow{3}{*}{None} & 32 & 109\x 10\x 3 & 33 & 77.8 & 92.8 \\
SlowFast~\cite{slowfast} &  R101+NL &   & 16+64 & 234\x10\x3 & 60 & 79.8 & 93.9 \\
MViTv1~\cite{mvit} & MViTv1-B &   & 32 & 170\x5\x1 & 37  & 80.2 & 94.4 \\
\hline
MAE-ST$_{1600e}$~\cite{feichtenhofer2022maest} & ViT-B & \multirow{3}{*}{Kinetics-400 w/o label} & 16 & 180\x7\x3 & 87 & 81.3 & 94.9 \\
VideoMAE$_{800e}$~\cite{tong2022videomae} & ViT-B &  & 16 & 180\x5\x3 & 87 & 80.0 & 94.4 \\
VideoMAE$_{1600e}$~\cite{tong2022videomae} & ViT-B &  & 16 & 180\x5\x3 & 87 & 81.5 & \textbf{95.1} \\
\shline\hline
MGMAE$_{800e}$ & ViT-B & \multirow{2}{*}{Kinetics-400 w/o label} & 16 & 180\x5\x3 & 87 & 81.2 & 94.9 \\
\textbf{MGMAE$_{1600e}$} & ViT-B & & 16 & 180\x5\x3 & 87 & \textbf{81.8} & 95.0
\end{tabular}
\vspace{-0.5mm}
\caption{\textbf{Comparison on the Kinetics-400 dataset}. We only list the results obtained with the similar backbones.}
\vspace{-1em}
\label{tab:sota-k400}
\end{table*}
\begin{figure}[t]
    \centering
    \includegraphics[width=\linewidth]{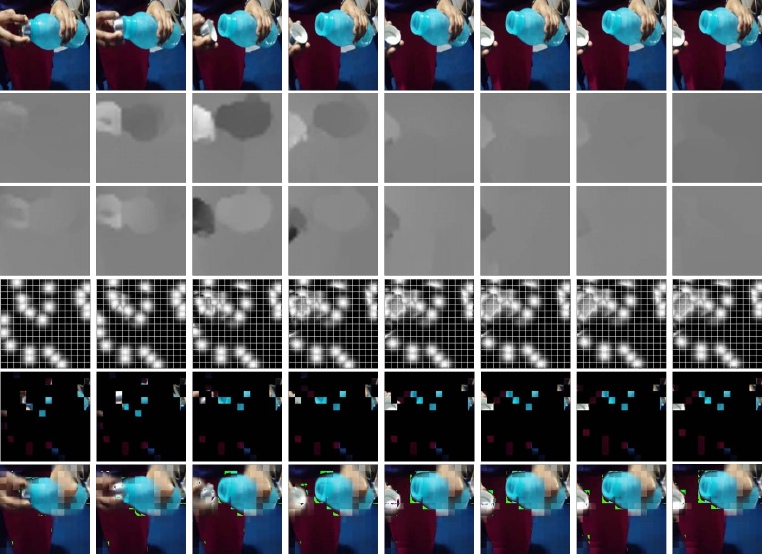}
    \caption{Visualization of the video in SSV2 validation set. We present our motion guided mask maps and the reconstructed images. From top to bottom, the original images, the x-direction flows, the y-direction flows, the motion guided mask maps, the masked images and the reconstructed images.}
    \label{fig:visualization}
\end{figure}

\paragraph{Pre-train loss implies a more challenging task.} The core design of MGMAE is to dynamically sample the positions of masked token under the guidance of optical flows and aims at increasing the difficulty of the reconstruction task. As can be seen in \cref{tab:pretrain_loss}, the pre-train loss of MGMAE is always larger than that of VideoMAE by more than 0.05. This loss gap implies motion guided masking further suppresses information leakage and indeed constructs a more challenging mask-reconstruct pretext task for masked video modeling. This more difficult task would like to encourage learning more effective representations.

\paragraph{Detailed breakdown of comparison between MGMAE and VideoMAE.} To understand the distinct impacts of the MGMAE and VideoMAE masking strategies on video model pre-training, we delved deeper into the per-class accuracy variations between the two. \cref{fig:acc_vs} showcases the 29 categories with the most pronounced differences in classification accuracy between MGMAE and VideoMAE.

\paragraph{MGMAE: a more effective video representation learner.} MGMAE benefits greatly from the harder task constructed by our motion guided masking strategy. On the one hand, the model has to encode the relationship between visible and invisible tokens harder, which can better guide the model training. On the other hand, the suppression of information leakage may well reduce the overfitting risk of the pre-training, and thus the model can be pre-trained much longer. As shown in \cref{tab:performance_compare}, MGMAE consistently maintains an obvious fine-tuning performance gap with VideoMAE on the motion-centric SSV2 dataset (1.4\% at 800 epochs and 1.5\% at 2400 epochs) and also has some improvement on the scene-centric Kinetics-400 dataset (1.2\% at 800 epochs and 0.3\% at 1600 epochs).

\paragraph{Visualization.}
We randomly seletec a video clip in SSV2 validation set and show its reconstruction example in \cref{fig:visualization}. We can see that the mask map changes with object movements, which makes it more difficult for the model to reconstruct the original video.

\subsection{Comparison with the state-of-the-art methods}
We compare our approach with the previous state-of-the-art methods on the Kinetics-400 and Something-Something V2 datasets. The results are shown in Table~\ref{tab:sota-k400} and Table~\ref{tab:sota-ssv2}. For a fair comparison, we mainly list the results with similar computational cost. On the Something-Something V2 dataset, our MGMAE with ViT-B backbone achieves a performance of 72.3\% when trained for 2400 epochs, which outperforms the original VideoMAE by 1.5\%. On the Kinetics-400 dataset, our MGMAE obtains slightly better performance than the original VideoMAE. The small performance improvement might be ascribed to the fact that Kinetics-400 is a scene-centric action recognition benchmark and motion information is less important compared with the Something-Something dataset. 

\section{Conclusion}
In this paper, we have proposed the Motion Guided Masked Autoencoders (MGMAE), which adapts the motion guided masking strategy to dynamically sample the unmasked visible tokens under the guidance of flows, thus suppressing information leakage to build a more challenging task for masked video pre-training. Experiments have shown that MGMAE has good performance and maintains a high-performance advantage over the previous methods under fair comparison. In addition, our strategy also reduces the risk of pre-training overfitting, which allows the model to benefit from longer pre-training.

\paragraph {\bf Acknowledgements.} {This work is supported by the National Key R$\&$D Program of China (No. 2022ZD0160900, No.2022ZD0160100), the National Natural Science Foundation of China (No. 62076119, No. 61921006), Shanghai Committee of Science and Technology (Grant No. 21DZ1100100), and Collaborative Innovation Center of Novel Software Technology and Industrialization.}

{\small
\bibliographystyle{ieee_fullname}
\bibliography{mgmae}
}

\end{document}